%% file: my_submission_final.tex
\documentclass{article}
\usepackage{spconf}
\usepackage{cite}
\usepackage{amssymb,amsfonts}
\usepackage[fleqn]{amsmath}
\usepackage{algorithmic}
\usepackage{graphicx}
\usepackage{textcomp}
\usepackage{xcolor}
\usepackage{booktabs}
\usepackage{multirow}
\usepackage{graphicx}
\usepackage{hyperref}

\usepackage{enumitem}
\setlist{nosep, leftmargin=14pt}

\usepackage{mwe} 

\title{Improving Normative Modeling for Multi-modal Neuroimaging Data using mixture-of-product-of-experts variational autoencoders}
\name{Sayantan Kumar$^{1,2}$ \qquad Philip Payne$^{1,2}$ \qquad Aristeidis Sotiras$^{2,3}$}

\address{$^1$ Department of Computer Science and Engineering, Washington University in St. Louis, USA \\
$^2$ Institute for Informatics, Data Science and Biostatistics, Washington University in St.Louis, USA \\ $^3$ Department of Radiology, Washington University in St.Louis, USA} 

\begin{document}

\maketitle

\begin{abstract}

Normative models in neuroimaging learn the brain patterns of healthy population distribution and estimate how disease subjects like Alzheimer's Disease (AD) deviate from the norm. Existing variational autoencoder (VAE)-based normative models using multimodal neuroimaging data aggregate information from multiple modalities by estimating product or averaging of unimodal latent posteriors. This can often lead to uninformative joint latent distributions which affects the estimation of subject-level deviations. In this work, we addressed the prior limitations by adopting the Mixture-of-Product-of-Experts (MoPoE) technique which allows better modelling of the joint latent posterior. Our model labelled subjects as outliers by calculating deviations from the multimodal latent space. Further, we identified which latent dimensions and brain regions were associated with abnormal deviations due to AD pathology.
\end{abstract}
\vspace{-2mm}

\section{Introduction}
\vspace{-1mm}

\noindent Normative modelling for neurodegenerative disorders like Alzheimer's Disease (AD) learn the brain patterns of cognitively unimpaired (healthy control) subjects. The trained model can be used to detect deviations at subject-level related to AD pathology \cite{dong2017heterogeneity, marquand2019conceptualizing}. Recent works on VAE-based normative models have a unimodal structure with a single encoder and decoder that can handle only a single imaging modality \cite{kumar2023normative, lawry2023multi, pinaya2019using, pinaya2021using}. However, brain disorders like AD are multifactorial, showing deviations from the norm in features across multiple imaging modalities. Since each modality is sensitive to disease effects to a varying degree, it is important to develop normative models that can handle multiple imaging modalities (e.g. Magnetic Resonance Imaging (MRI) and Positron Emission Tomography (PET)).

Existing multimodal VAE frameworks for normative modelling have modality-specific encoders and decoders and aggregate the encoding distributions to learn a joint latent representation \cite{lawry2023multi}. Each modality can be treated as an 'expert' and the aggregated latent distributions are estimated by either taking the product (Product-of-Experts (PoE) \cite{wu2018multimodal}) or summation across the unimodal expert's densities (Mixture-of-Experts (MoE) \cite{shi2019variational}). However, both PoE and MoE have their own sets of limitations. PoE joint distribution may be biased towards overconfident but miscalibrated experts, ignoring other modalities and this leads to a sub-optimal joint representation \cite{shi2019variational}. In MoE, the aggregating experts do not result in a distribution that is sharper than the other experts. Even if we increase the number of experts, the shared latent representation does not become more informative as in PoE, which means that no proper aggregated inference is possible in the latent space.

To address the above limitations of MoE and PoE, we adopted the Mixture-of-Product-of-Experts (MoPoE) approach \cite{sutter2021generalized} for aggregating information from multiple modalities in the latent space. MoPoE combines the strengths of both MoE and PoE, leading to more informative joint latent distribution. Previous works have shown that normative deviations calculated in the latent space outperform feature-space deviations \cite{lawry2022conditional,lawry2023multi}. Our model quantified the deviation of AD patients based on the joint latent space and label subjects with abnormal (statistically significant) deviations as outliers. We showed that normative modelling with MoPoE better captures outliers compared to the baseline methods. We clinically validated the latent deviations to check if they were sensitive to the different AD stages and if they were significantly associated with cognition. Finally, we identified latent dimensions with abnormal deviations, mapped them to feature-space deviations and analyzed regional brain deviations associated with AD pathology.

\vspace{-2.5mm}


\setlength{\mathindent}{0pt}
\section{Methods}

\vspace{-1mm}

\subsection{Approximate the joint posterior}

\noindent Let $X = [x_1, x_2,...x_N]$ be a set of conditionally independent N modalities. The backbone of our model is a multimodal variational autoencoder (mVAE), a generative model of the form $p_\theta(x_1, x_2,...x_N, z) = p(z)\prod_{i=1}^{N}p_\theta(x_i|z)$, where z is the latent variable and $p(z)$ is the prior. mVAE optimizes the ELBO (Evidence Lower Bound) which is a combination of modality-specific likelihood distributions $p_\theta\left(x_i|z\right)$ and the KL divergence between the approximate joint posterior $q(z|X)$ and prior $p(z)$. 

\vspace{-3mm}
\begin{equation*}
\operatorname{ELBO} = \mathbb{E}_{q(z \mid X)}\left[\sum_{x_i \in X} \log p_\theta\left(x_i|z\right)\right]-\operatorname{KL}\left[q_\phi(z|X),p(z)\right]
\end{equation*}
\vspace{-3mm}

\noindent In mVAE, each modality can be treated as an "expert" and the approximate joint posterior can be estimated either by taking the product of the unimodal posteriors (Product-of-Experts (PoE))  $q_{PoE}(z|X) = p(z) \prod_{i = 1}^{N} q_\phi(z|x_i)$ \cite{wu2018multimodal} or by the Mixture-of-Experts (MoE) approach, where the joint inference distribution is represented by the sum of the unimodal inferences $q_{MoE}(z|X)=\frac{1}{N} \sum_{i = 1}^{N} q_{\phi_i}(z|x_i)$ \cite{shi2019variational}. In PoE, if the inference for a particular modality is very sharp, the joint inference will be heavily dominated by it. Hence, the optimization of unimodal inference with low precision might be greatly degraded. On the other hand, MoE spreads its density over all individual experts and hence the joint latent posterior is not sharper than the unimodal posteriors \cite{daunhawer2021limitations}.

\begin{figure}[!htbp]
    \centering
    \includegraphics[width = \linewidth]{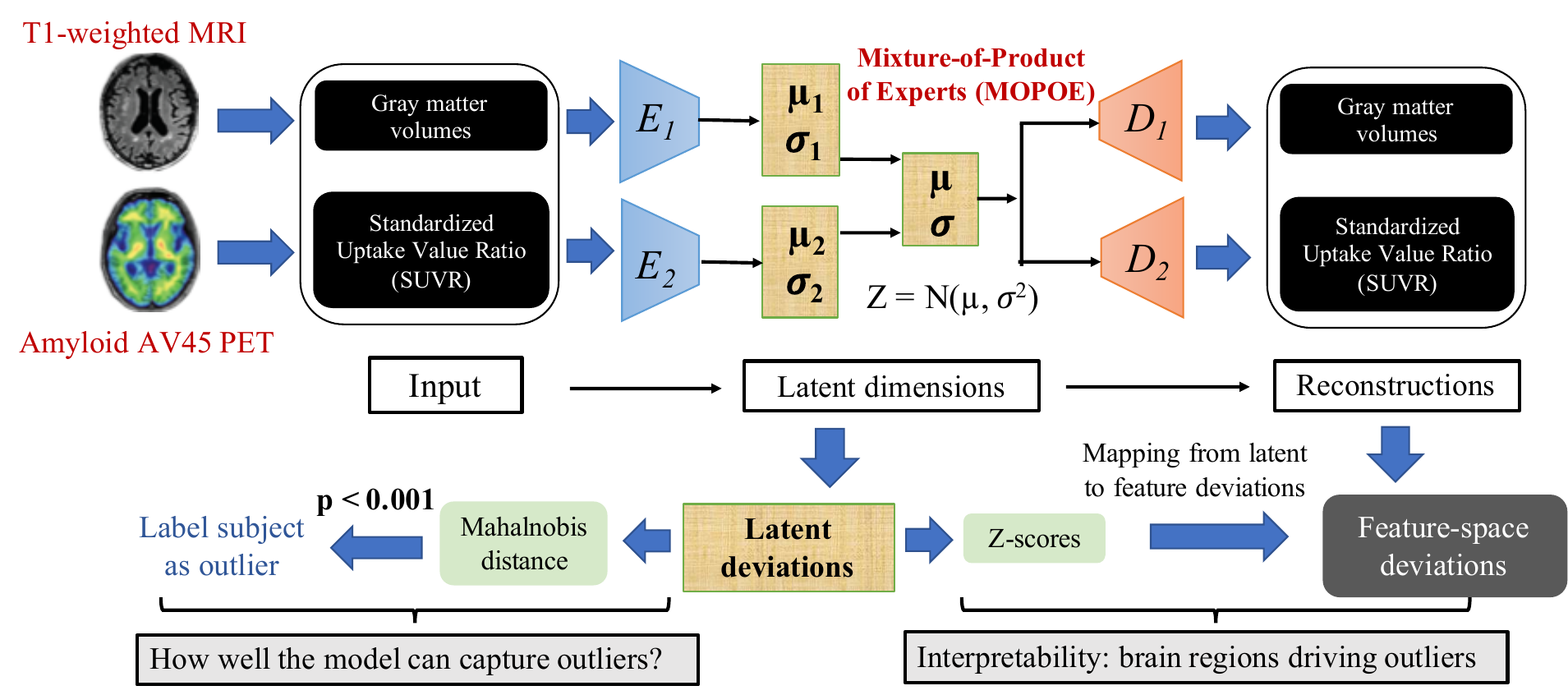}
    \vspace{-10pt}
    \caption{Proposed MoPoE normative modelling framework}
    \vspace{-10pt}
    \label{fig:workflow}
\end{figure}

\vspace{-2mm}
\subsection{Mixture-of-Product-of-Experts}

\noindent To mitigate the challenges of both PoE and MoE, we adopted a generalization of both MoE and PoE, called Mixture-of-Product-of-Experts (MoPoE) proposed by \cite{sutter2021generalized} (Equation 2) where $X_k$ represents a random subset of N modalities and $P(X)$ represents the power-set of all N modalities. MoPoE can be considered as a hierarchical distribution. First, the  the unimodal posterior approximations of a subset $X_k$ are combined by PoE (Equation 1). Next, the subset approximations $q_\phi(z \mid X_k)$ are combined by MoE (Equation 2). This combines the strengths of both MoE and PoE while addressing their weaknesses. The corresponding ELBO for MoPoE is represented by Equation 3.

\vspace{-5mm}
\begin{gather}
q_{PoE}(z \mid X_k) \propto \prod_{x_i \in X_k} q(z|x_i) \\
q_{MoPoE}(z|X) = \frac{1}{2^N} \sum_{X_k \in P(X)} q_{PoE}(z|X_k) \\
L = \mathbb{E}_{q(z \mid X)}\left[\sum_{x_i \in X} \log p_\theta\left(x_i|z\right)\right]- \operatorname{KL}\left[q_{MoPoE}(z|X),p(z)\right]
\end{gather}
\vspace{-10mm}

\subsection{Multi-modal normative modelling}

Our framework (Figure 1) has separate encoders to encode the 2 modalities into their corresponding latent parameters (mean and variance). The unimodal latents were aggregated through the MoPoE approach to estimate the shared latent parameters (joint latent distribution). The shared latents were passed through the modality-specific decoders to reconstruct each modality. The model was first trained to characterize the healthy population cohort. We assumed disease abnormality can be quantified by measuring how AD subjects deviate from the joint space (latent deviations) \cite{lawry2022conditional, lawry2023multi} or from the reconstruction errors of healthy controls (feature-space deviations) \cite{kumar2023normative, pinaya2021using}. At test time, the trained model was applied on the AD cohort to estimate both latent and feature deviations.  

\vspace{-2mm}
\subsection{Multi-modal latent and feature deviations}

\noindent \textbf{Mahalanobis distance}: To quantify how much AD each subject deviates from the latent distribution of healthy controls, we measured the Mahalanobis distance \cite{lawry2023multi}, which accounts for correlations between latent vectors.

\vspace{-5mm}
\begin{align*}
D_{ml} &=\sqrt{(z_j - \mu(z_{norm}))^\top\Sigma(z_{norm})^{-1}(z_j - \mu(z_{norm}))} 
\end{align*}
\vspace{-5mm}

\noindent where $z_j \equiv q(z_j|X_j)$ is a sample from the joint posterior distribution for subject $j$. $\mu(z_{norm})$ and $\Sigma(z_{norm})$ represent the mean and covariance of the healthy cohort latent position. We additionally derived a multivariate feature-space deviation index based on the Mahalonobis distance that quantifies how the reconstruction error of an AD subject deviate from the reconstruction errors of controls.

\vspace{-5mm}
\begin{align*}
D_{mf} &=\sqrt{(d_j - \mu(d_{norm}))^\top\Sigma(d_{norm})^{-1}(d_j - \mu(d_{norm}))} 
\end{align*}
\vspace{-5mm}

\noindent where $d_j = [d^1_j, ...d^i_j, ..d^R_j]$ is the mean squared reconstruction error between original and reconstructed input for subject j and brain region $i = [1, 2,..R]$. $\mu(d_{norm})$ and $\sigma(d_{norm})$ are the mean and covariance of the healthy cohort reconstruction error respectively. 

\vspace{3pt}

\noindent \textbf{Z-score latent deviations}: To analyse which latent dimensions and brain regios are associated with abnormal deviations due to AD pathology, we calculated both latent space Z-scores $Z_{ml}$ (for each latent dimension) and feature-space Z-scores $Z_{mf}$. $z_{ij}$ and $d_{ij}$ represent the latent values and reconstruction error of test subject j for i-th position respectively. $\mu(z_{ij}^{norm})$ ($\mu(d_{ij}^{norm})$) and $\sigma(z_{ij}^{norm})$ ($\sigma(d_{ij}^{norm})$) are the mean and standard deviations of healthy cohort latent values (reconstruction error) respectively. 

\vspace{-5mm}
\begin{align*}
Z_{ml} = \frac{z_{ij} - \mu(z_{ij}^{norm})}{\sigma(z_{ij}^{norm})} && Z_{mf} = \frac{d_{ij} - \mu(d_{ij}^{norm})}{\sigma(d_{ij}^{norm})}
\end{align*}
\vspace{-5mm}


\input{Plots/sig_ratio_table}

\section{Experiments and Results}
\vspace{-1mm}
\subsection{Data and feature processing}

\noindent For training, we selected 248 cognitively unimpaired (healthy control) subjects from the Alzheimer’s Disease Neuroimaging (ADNI) dataset \cite{mueller2005ways} with Clinical Dementia Rating (CDR) = 0 and no amyloid pathology. We used regional brain volumes extracted from T1-weighted MRI scans and regional Standardized Uptake Value Ratio (SUVR) values extracted from AV45 Amyloid PET scans as the two input modalities for our model (Figure 1). Both brain volumes and SUVR values were extracted from 66 cortical (Desikan-Killiany atlas) and 24 subcortical regions. At test time, we used 48 healthy controls for a separate holdout cohort and a disease cohort of 726 individuals across the following AD stages: preclinical stage with no symptoms (CDR = 0, A+) (N = 305), (b) CDR = 0.5 (N = 236) and (c) $CDR >= 1 (N = 185)$. Each brain ROI was normalised by removing the mean and dividing by the standard deviation of the healthy control cohort brain regions. We conditioned our model on the age and sex of patients, represented as one-hot encoding vectors, to remove the effects of covariates from the MRI and PET features.

\vspace{-2mm}
\subsection{Baselines and implementation details}

\noindent We compared our proposed MoPoE framework to the following classes of baselines: (i) \textbf{aggregation strategies in normative modelling}: MoE and generalized PoE \cite{lawry2023multi}, PoE \cite{kumar2023normative}, (ii) \textbf{unimodal baselines}: MRI only, amyloid only and concatenated MRI + amyloid and (iii) \textbf{state-of-the-art multimodal VAE models}: mmJSD \cite{sutter2020multimodal}, JMVAE \cite{suzuki2016joint} and MVTCAE \cite{hwang2021multi}. All baseline models except the unimodal ones were implemented using the open-source Multi-view-AE python package \cite{aguila2023multi}. All models were trained using Adam optimizer with hyperparameters as follows: epochs = 500, learning rate = $10^{-5}$, batch size = $64$ and latent dimensions in the range [5,10,15,20]. The encoder and decoder networks have 2 fully-connected layers of sizes ${64, 32}$ and ${32, 64}$ respectively.

\vspace{-2mm}
\subsection{Evaluating outlier detection performance}

\noindent For each model, we calculated $D_{ml}$ and $D_{mf}$ for the healthy holdout cohort and disease cohort from ADNI. For both deviation metrics, we identified subjects with statistically significant ($p<0.001$) deviations \cite{tabachnick2013using}. A good normative model is supposed to correctly identify disease subjects as outliers and healthy individuals lying within the normative distribution. We used the positive likelihood ratio for assessing how well our MoPoE normative model can detect outliers.

\vspace{-5mm}
\begin{equation*}
\text{likelihood ratio} = \frac{\text{True positive rate}}{\text{False positive rate}} = 
\frac{\frac{N_{disease}(outliers)}{N_{disease}}}{\frac{N_{holdout}(outliers)}{N_{holdout}}}
\end{equation*}
\vspace{-2mm}

\noindent Latent deviations $D_{ml}$ achieved greater likelihood ratios compared to feature-space deviations $D_{mf}$ (Table 1). All models performed similarly when using $D_{mf}$. Our proposed model with MoPoE achieved the best overall performance across different latent dimensions with the highest likelihood ratio for d = 5 and d = 10. Also, the multimodal models performed better than than the unimodal ones, indicating that better normative models can be learnt by modelling the joint distribution between modalities.

\begin{figure}[htbp]
    \centering
    \includegraphics[width = \linewidth]{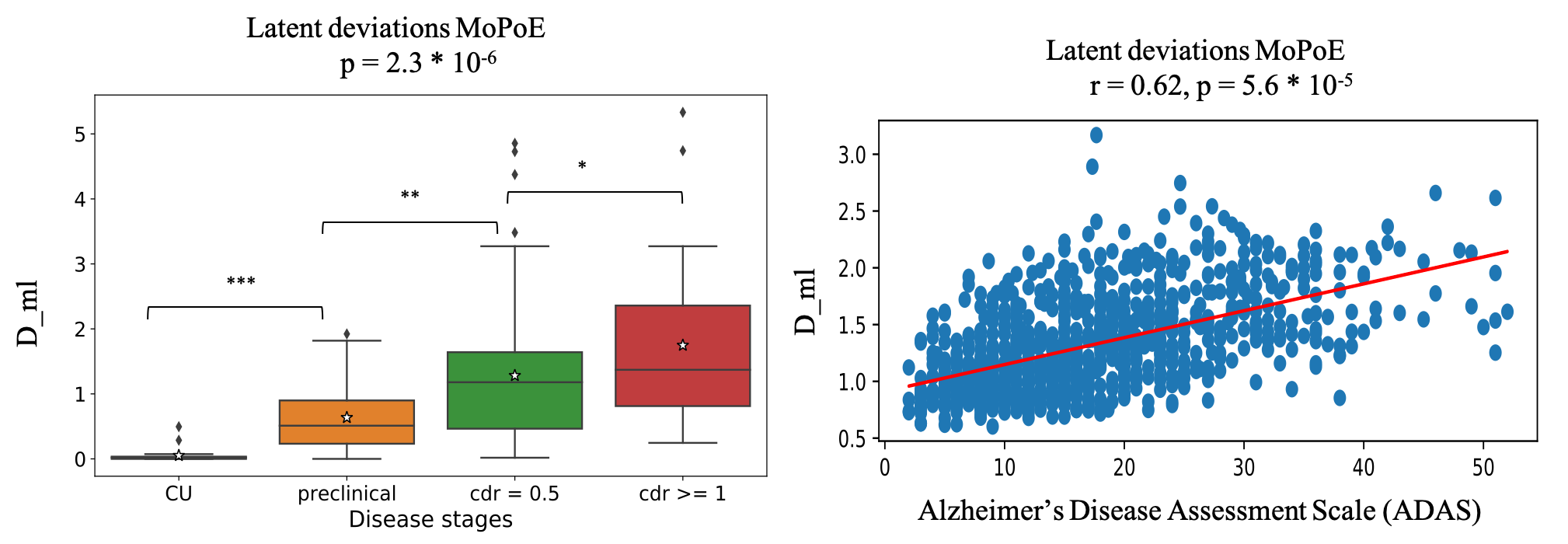}
    \vspace{-10pt}
    \caption{\textbf{Left}: Box plot showing the latent deviations $D_{ml}$ across cognitively unimpaired (CU) subjects and the AD groups (in order of severity). \textbf{Statistical annotations}: ns: not significant, $0.05 < p <= 1$: *,$0.01 < p <= 0.05$: **, $0.001 < p < 0.01$: ***, $p < 0.001$. \textbf{Right}: Association between $D_{ml}$ and cognition scores (ADAS). Each point in the plot represents a subject and the red line denotes the linear regression fit of the points, adjusted by age and sex.}
    \vspace{-10pt}
    \label{fig:workflow}
\end{figure}

\begin{figure*}[htbp]
    \centering
    \includegraphics[width = 0.75\linewidth]{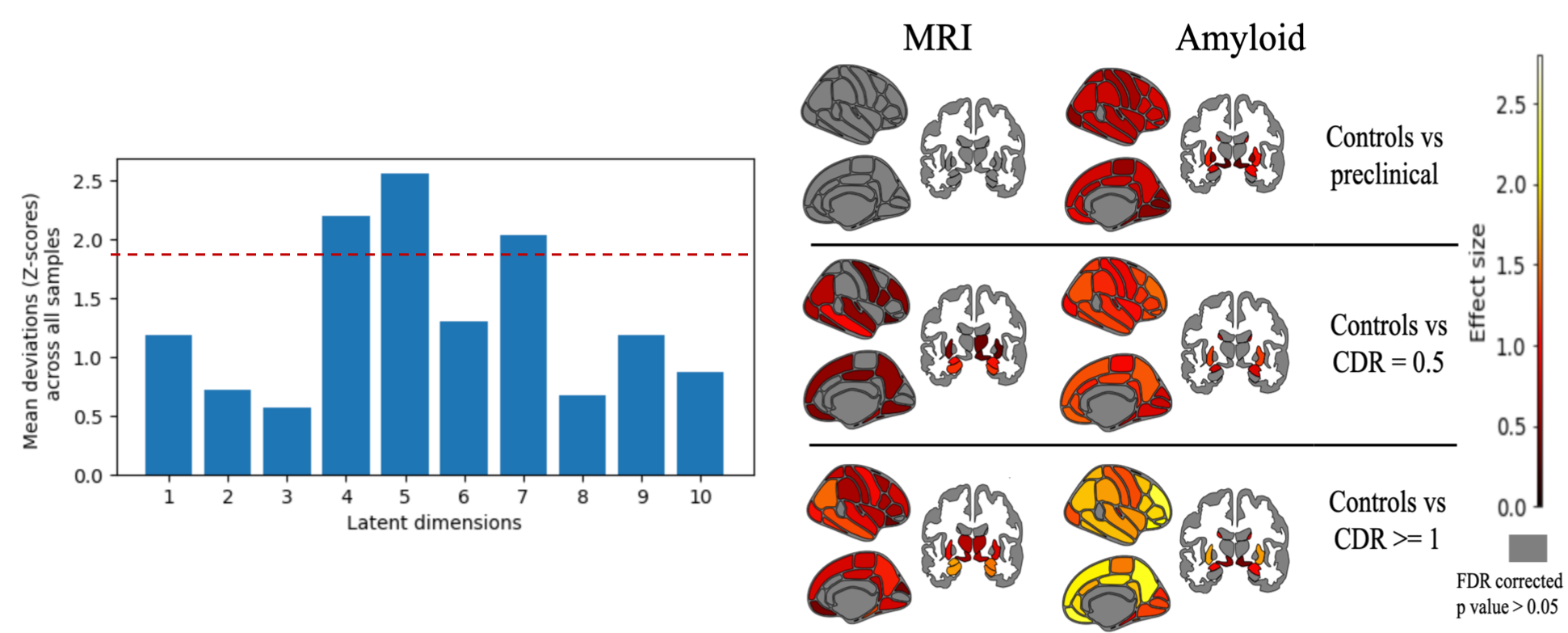}
    \vspace{-10pt}
    \caption{\textbf{Left}: Latent dimensions (4,5 and 7) with statistically significant deviations (mean absolute $Z_{ml} > 1.96$ or $p<0.05$). The dotted red line indicates $Z>1.96$. Latent dimensions above the dotted line were used for mapping to feature-space deviations. \textbf{Right}: Effect size maps showing the region-level pairwise group differences in $Z_{mf}$ between control subjects and each of the AD stages for both the modalities. The color bar represents the Cohen’s d statistic effect size (0.5 is considered a small effect, 1.5 a medium effect and 2.5 a large effect). Gray regions represent that no participants have statistically significant deviations after False Discovery Rate (FDR) correction.}
    \vspace{-10pt}
    \label{fig:workflow}
\end{figure*}

\vspace{-2mm}
\subsection{Clinical validation of latent deviations}

\noindent We demonstrated the clinical validation of our proposed multimodal latent deviations $D_{ml}$ generated by MoPoE in two ways: (i) sensitivity with respect to disease staging and (ii) correlation with subject cognition. For the ADNI dataset, $D_{ml}$ showed a monotonous increasing pattern across the AD stages and differed between groups overall (\textbf{Figure 2 left}). Thus, latent deviations generated by our proposed model can capture the neuroanatomical alterations in the brain due to the progressive stages of AD. 

\noindent We also measured the association between $D_{ml}$ and the Alzheimer's Disease Assessment Scale (ADAS) cognition score \cite{rosen1984new}. ADAS scores range from 0 to 70 with high scores indicating worse cognition. Age and sex-adjusted linear regressions were fitted between the $D_{ml}$ and cognition scores, with Pearson Correlation Coefficient measuring the correlation between them. Our results showed that latent deviations were significantly associated with ADAS scores ($\beta = 1.2$, $p = 5.6 * 10^{-5}$) with a correlation coefficient of 0.62 (\textbf{Figure 2 right}).

\vspace{-2mm}
\subsection{Interpretability analysis}

\subsubsection{Mapping from latent to feature deviations}

Ideally, all latent dimensions can be used to reconstruct the input data and quantify feature-space deviations. The latent dimensions with statistically significant ($p<0.05$) mean absolute Z-scores $Z_{ml}$ indicate the latent dimensions which show deviation between control and disease cohorts. This can provide an interpretation of how latent space deviations can be mapped to deviations in the feature-space. We passed these selected latent vectors through the decoders setting the remaining latent dimensions and covariates to be 0 such that the reconstructions and feature deviations $Z_{mf}$ reflect only the information encoded in the selected latent vectors. From the ADNI dataset, we identified 3 out of 10 latent dimensions (4,5 and 7) whose mean absolute $Z_{ml} > 1.96$ (p < 0.05) and used them for generating the feature-space deviations $Z_{mf}$ (\textbf{Figure 3 left}).

\vspace{-3mm}
\subsubsection{Brain regions associated with AD abnormality}

We visualized pairwise group comparisons in $Z_{mf}$ for each modality at each of 90 regions using the Cohen’s d-statistic effect size maps after False Discovery Rate (FDR) correction, adjusted by age and sex. High effect size corresponding to a region indicates significantly more grey matter volume atrophy (neurodegeneration) or amyloid deposition (pathological abnormalities) in the brain compared to healthy control subjects. Region-level group differences for MRI atrophy were most evident within the frontal, temporal and hippocampal and amygdala regions with maximum effect size in the amygdala and hippocampal regions (Figure 2B). For amyloid, the differences extended to all cortical regions, ventricles, and the amygdala with the accumbens area region showing the maximum effect size (\textbf{Figure 3 right}). 


\vspace{-2mm}
\section{Conclusion}

In this work, we built on recent studies \cite{lawry2023multi, kumar2023normative} and introduced a multimodal VAE normative model which provides an alternative method of learning the joint normative distribution between multiple modalities to address the limitations of existing approaches. We adopted the MoPoE technique of aggregating information from modalities which results in more informative latent space and better capture of outliers as evidenced by the better significance ratio for the ADNI dataset. The latent deviations were found to be sensitive towards the AD stages and significantly associated with subject cognition. We also identified latent dimensions with statistically Z-score deviations, mapped only those selected latent vectors to feature-space deviations and analyzed brain regions associated with AD abnormality for both modalities. 

\vspace{-2mm}
\section{Compliance with Ethical Standards}

Ethical approval was not required as confirmed by the license
attached with the open access data. The study was approved by the Institutional Review Board at the Washington University in St. Louis.

\vspace{-2mm}
\section{Acknowledgement}

\noindent The preparation of this report was supported by the Centene Corporation contract (P19-00559) for the Washington University-Centene ARCH Personalized Medicine Initiative and the National Institutes of Health (NIH) (R01-AG067103). Computations were performed using the facilities of the Washington University Research Computing and Informatics Facility, which were partially funded by NIH grants S10OD025200, 1S10RR022984-01A1 and 1S10OD018091-01. Additional support is provided The McDonnell Center for Systems Neuroscience.

\vspace{5pt}

\noindent Data used in preparation of this article were obtained from the Alzheimer’s disease Neuroimaging Initiative (ADNI) database (http://adni.loni.usc.edu). As such, the investigators within the ADNI contributed to the design and implementation of ADNI and/or provided data but did not participate in analysis or writing of this report. A complete listing of ADNI investigators can be found \href{http://adni.loni.usc.edu/wp-content/uploads/how_to_apply/ADNI_Acknowledgement_List.pdf}{here}

\vspace{-2mm}
\bibliographystyle{IEEEbib}
\bibliography{references}

\end{document}

%% file: Plots/sig_ratio_table.tex
\begin{table}[]
\centering
\caption{Likelihood ratio calculated for $D_{ml}$, $D_{mf}$ (ADNI)}
\label{tab:my-table}
\resizebox{\columnwidth}{!}{%
\begin{tabular}{|l|cccc|cccc|}
\hline
\multicolumn{1}{|c|}{\multirow{2}{*}{Latent dimensions}} &
  \multicolumn{4}{c|}{Latent Mahalnobis} &
  \multicolumn{4}{c|}{Feature Mahalnobis} \\ \cline{2-9} 
\multicolumn{1}{|c|}{} &
  \multicolumn{1}{c|}{d = 5} &
  \multicolumn{1}{c|}{d = 10} &
  \multicolumn{1}{c|}{d = 15} &
  d = 20 &
  \multicolumn{1}{c|}{d = 5} &
  \multicolumn{1}{c|}{d = 10} &
  \multicolumn{1}{c|}{d = 15} &
  d = 20 \\ \hline
MOPOE (proposed) &
  \multicolumn{1}{c|}{\textbf{6.73}} &
  \multicolumn{1}{c|}{\textbf{8.1}} &
  \multicolumn{1}{c|}{5.72} &
  4.56 &
  \multicolumn{1}{c|}{1.85} &
  \multicolumn{1}{c|}{2.52} &
  \multicolumn{1}{c|}{2.16} &
  1.76 \\ \hline
POE &
  \multicolumn{1}{c|}{6.21} &
  \multicolumn{1}{c|}{7.95} &
  \multicolumn{1}{c|}{5.68} &
  4.89 &
  \multicolumn{1}{c|}{1.67} &
  \multicolumn{1}{c|}{1.92} &
  \multicolumn{1}{c|}{2.22} &
  1.32 \\ \hline
MOE &
  \multicolumn{1}{c|}{6.26} &
  \multicolumn{1}{c|}{7.57} &
  \multicolumn{1}{c|}{5.45} &
  \textbf{5.68} &
  \multicolumn{1}{c|}{1.37} &
  \multicolumn{1}{c|}{1.53} &
  \multicolumn{1}{c|}{1.77} &
  1.68 \\ \hline
gPOE &
  \multicolumn{1}{c|}{6.64} &
  \multicolumn{1}{c|}{7.8} &
  \multicolumn{1}{c|}{\textbf{5.75}} &
  4.3 &
  \multicolumn{1}{c|}{2.11} &
  \multicolumn{1}{c|}{1.65} &
  \multicolumn{1}{c|}{1.86} &
  1.55 \\ \hline
mri only &
  \multicolumn{1}{c|}{4.92} &
  \multicolumn{1}{c|}{4.65} &
  \multicolumn{1}{c|}{4.25} &
  4.67 &
  \multicolumn{1}{c|}{-} &
  \multicolumn{1}{c|}{-} &
  \multicolumn{1}{c|}{-} &
  - \\ \hline
amyloid only &
  \multicolumn{1}{c|}{5.25} &
  \multicolumn{1}{c|}{5.12} &
  \multicolumn{1}{c|}{4.41} &
  4.83 &
  \multicolumn{1}{c|}{-} &
  \multicolumn{1}{c|}{-} &
  \multicolumn{1}{c|}{-} &
  - \\ \hline
mri\_amyloid\_concat &
  \multicolumn{1}{c|}{4.72} &
  \multicolumn{1}{c|}{5.95} &
  \multicolumn{1}{c|}{4.43} &
  4.2 &
  \multicolumn{1}{c|}{1.42} &
  \multicolumn{1}{c|}{1.67} &
  \multicolumn{1}{c|}{1.33} &
  1.2 \\ \hline
mmJSD &
  \multicolumn{1}{c|}{6.31} &
  \multicolumn{1}{c|}{7.21} &
  \multicolumn{1}{c|}{4.66} &
  5.25 &
  \multicolumn{1}{c|}{2.21} &
  \multicolumn{1}{c|}{1.81} &
  \multicolumn{1}{c|}{1.4} &
  1.25 \\ \hline
JMVAE &
  \multicolumn{1}{c|}{5.68} &
  \multicolumn{1}{c|}{7.85} &
  \multicolumn{1}{c|}{5.1} &
  4.51 &
  \multicolumn{1}{c|}{1.68} &
  \multicolumn{1}{c|}{2.33} &
  \multicolumn{1}{c|}{2.1} &
  1.75 \\ \hline
MVTCAE &
  \multicolumn{1}{c|}{6.34} &
  \multicolumn{1}{c|}{6.83} &
  \multicolumn{1}{c|}{5.56} &
  3.82 &
  \multicolumn{1}{c|}{1.35} &
  \multicolumn{1}{c|}{1.68} &
  \multicolumn{1}{c|}{1.52} &
  1.21 \\ \hline
\end{tabular}%
}
\end{table}